# Evolution of SLAM: Toward the Robust-Perception of Autonomy

B. Udugama, MSc Robotics Student, Middlesex University

*Abstract*—Simultaneous localisation and mapping (SLAM) is the problem of autonomous robots to construct or update a map of an undetermined unstructured environment while simultaneously estimate the pose in it. The current trend towards self-driving vehicles has influenced the development of robust SLAM techniques over the last 30 years. This problem is addressed by using a standard sensor or a sensor array (Ultrasonic sensor, LIDAR, Camera, Kinect RGB-D) with sensor fusion techniques to achieve the perception step. Sensing method is determined by considering the specifications of the environment to extract the features. Then the usage of classical Filter-based approaches, the global optimisation approach which is a popular method for visual-based SLAM and convolutional neural network-based methods such as deep learning-based SLAM are discussed whereas considering how to overcome the localisation and mapping issues. The robustness and scalability in long-term autonomy, performance and other new directions in the algorithms compared with each other to sort out. This paper is looking at the published previous work with a judgemental perspective from sensors to algorithm development while discussing open challenges and new research frontiers.

## I. INTRODUCTION

CAPABILITY to manoeuvre over a complicated unknown surrounding is one of the foremost necessary challenges in the field of autonomous robotics [1]. The solution for this problem was divided into two frontiers: (1) the Simultaneous Localization and Mapping (SLAM) that will generate the map information while localising the robot in the environment, and (2) the Navigation algorithm that describes the traverse path towards the goal while avoiding obstacles [2], [3], [5]. The success of the Autonomous robotics field is hugely dependent upon the solution of this SLAM problem and will have various applications ranging from agriculture to oil exploration, medical applications to nuclear laboratories, and intelligent vehicles to spatial expeditions [4]. The research focus on this topic has increased with the aid of the automobile industry due to the current trend towards self-driving cars.

Furthermore, for robust and exact estimation of the localisation, one could propose the utilisation of Global Navigation Satellite System (GNSS) which will provide autonomous geospatial information with global coverage to accomplish a noteworthy outcome. It will not be a perfect solution to this problem due to its drawbacks. Even though the precision of the traditional GNSS arrangements is raised with the usage of correctly aligned base stations, accessibility to such kind of a system in a global manner stays as a problem. Further GNSS is affected by spatial constraints that are unable to predict and eliminate. Especially the signal interference and line of sight disturbances could block the communication which has catastrophic outcomes with the erroneous localisation [6], [27].

Road markings or roadway identification can use to navigate a vehicle in a road such as Advanced-Driver-Assistance-Systems (ADAS) technology that is included with control for stability, new anti-locking, predictive cruise control and adaptive control of the pathway. This approach resolved the requirement of a distance communication system and focused on the local features to ensure the optimum and safe traversal in the road [11]. A significant drawback of these systems is that it needs a considerable number of identifiable characteristics such as in road signs and markings on the road. Increasingly complex conditions (urban for the most part, with convergences, bent streets, and so forth.) do not generally give enough data to confine a vehicle within a secure path. However, more information with precision and accuracy is required to develop a framework and guarantee reliable and safe navigation. Thus, various positioning systems ought to be considered.

The basic functionality of an autonomous vehicle is to localise itself with respect to a global or a local frame prior to any other planning or sensing steps [1]. Predicting the optimum safe path while evaluating other moving objects to avoid obstacles will be the next question to arise after obtaining its accurate position and orientation. The SLAM addresses this requirement, while yet being sufficiently general to permit the utilisation of any sensor or estimation procedure that suits the essential of evaluating both locating and mapping simultaneously. The mapping step is of prime intrigue when the self-driving is considered as it offers the first degree of discernment that is required to generate proper resolution of the navigation steps.

Solving the SLAM problem is treated as one of the major areas in autonomy and intrinsic part of self-driving robots [27]. Numerous issues are yet averting the utilisation of SLAM with autonomous vehicles that should travel for many kilometres in altogether different conditions. This elaborates the major problems for SLAM applications on autonomy: pose estimation tends to drift in long as well as maps are not feasible in almost any driving circumstance. The approximation of local and concurrent positioning provided by
maps are not feasible in almost any driving circumstance. The approximation of local and concurrent positioning provided by SLAM algorithms continues to deviate from the actual trajectory with the travelling distance [23]. Therefore, without prior knowledge or exact details, maintaining proper positioning during several kilometres becomes almost impossible. This brings us to the next problem, which is to provide maps that are appropriate for the task of localisation regardless of the conditions like the season, weather, geographical area, and traffic. Several approaches have been proposed to tackle this challenge, such as creating a map with careful monitoring of distinguishing data to reuse it later or use wireless communication infrastructures to exchange and improve the maps created by other road users [9], [12].

SLAM framework encompasses simultaneous evaluation of the state of a robot fitted with onboard sensors and the development of a map representation. Throughout this paper, the SLAM evolution from sensor data extraction to the algorithmic approach on state estimation and simultaneous mapping is discussed along with the loop closure technique which is used in the purpose of refining the localisation by revisiting the already mapped areas. However, different benchmarks and data sets available to experiment with algorithms is not discussed in this review.

## II. ANATOMY OF SLAM IMPLEMENTATIONS

A SLAM system consists of mainly two parts: called as the front-end and back-end implementations. Perception stage, which includes the sensing and modelling the data is the front end, while the back end carries out decision making based on the front end, generated abstract data. Exploration of these two areas is the focus of this section, beginning from the Maximum a Posteriori Estimation (MAP) [27], [29].

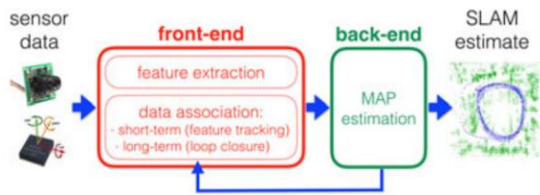

Fig. 1. SLAM process, with front end and back end. The back end provides feedback to the front end to track and validate the loop closure [27].

### A. MAP Estimation – SLAM Back-End

The new *de-facto* default SLAM definition came from Lu and Milios [57], which was preceded by Gutmann and Konolige [58]. From then on, most methods throughout the area of task analysis have increased efficiency and strength [7], [10], [27], [29]. All the above methods articulate SLAM as a top-down issue, and which is a maximum-posterior estimation problem [43] as a basis for the cooperation between states.

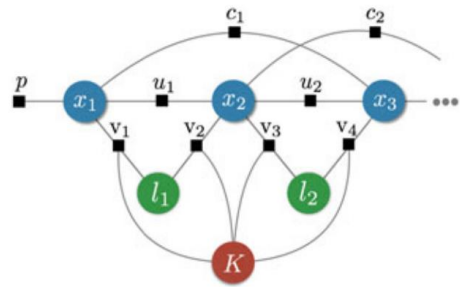

Fig. 2. *Graph representation of SLAM*: $(x_1, x_2, ...)$ represents the concurrent poses of the robot, $(l_1, l_2, ...)$ represent the landmark position estimations, $K$ represents the parameters for implicit calibration. Black squares are used to represent the state inputs: "u" describes the control inputs, "v" describes sensor observations, "c" describes loop closures, and "p" describes previous state factors [27].

### B. Perception-Based SLAM Front-End

In real robot implementations, as mandated by the MAP prediction, it would be challenging to read the sensor readings straight as an analytical state function. Of example, when the information of the unprocessed sensor is a raw image, the frequency of each pixel may be challenging to show based on the SLAM system, with more simplistic instruments. The same complexity occurs [19]. In each of these scenarios, it arises from the fact that we could not model a reasonably broad, but comprehensible, physical interpretation. It would be tough to write an analytical mechanism that connects the calculated variables even in the presence of this general representation.

Therefore, a front-end unit, which derives related unique features from the sensor information, is typically located before the SLAM back end. For example, in Visual SLAM, the front-end derives the pixel position of some distinct vital points in the world, which are now readily modelled in the back end. The front end often compares the different measurements in the field with a reference point: this is the so-called metadata association [27].

Figure 2 shows a visual depiction of a traditional SLAM process. A collection of shorter and longer-term information associations is included in the software association section at the front end. For shorter-term sensor quantification, the affiliation is accountable for the connection of co-responding features; for example, the shorter-term affiliation will monitor the fact that two-pixel measurements show the same 3-D point at a successive frame. At the other side, longer-term information correlation is responsible for the connection of modern quantities to earlier points of reference. The back end generally provides info back to the front end, e.g. for the identification of loop.

## III. SLAM FOR AUTONOMY I: ROBUSTNESS

In specific ways a SLAM framework may be error-prone: malfunction can depend on hardware or algorithmic approach. This former involves error modes precipitated by limitations of the established SLAM algorithms considering the difficulty in managing excessively diverse or challenging surroundings. Sensor deterioration or actuator destruction may cause the latter problem. For long-term deployment, the solution to such points of failure is critical because simplifying predictions about the ecosystem's configuration can no longer be established depending on the onboard sensors. The main obstacles to computational reliability are reviewed in this section and address open problems, including equipment-related vulnerabilities [29].

Data aggregation is one of the primary sources of algorithmic flaws. As stated in the Anatomy section for SLAM, that measurement follows the part of the state to which the information relates. In a graphical SLAM based on the feature, for instance, each visual interface is correlated with a function. This question is especially challenging due to the cognitive aliases, which results in a different sensory input resulting in the same sensor footprint. Data relations create inaccurate estimation-state matches in the involvement of perceptual associations. Besides, this results in false estimates of states. While, if the data relation improperly discards a sensor data as fraudulent (false negatives), calculating the precision of the measurements is affected due to lack of data. The condition is compounded by the existence of undescribed states of the dynamics, which can mislead the association of data. The latest SLAM approach takes a relatively general view that the environment remains consistent as the robot travels through it. The stationary universe hypothesis is valid in one map run in small-scale contexts if short-term movements are not present (e.g. moving individuals and objects). Change is necessary for long-term and widespread map levels.

The resiliency of SLAM in extreme environments such as the underwater is another aspect [14], [31]. The difficulties are the continually evolving conditions and the fact that it is tough to use traditional sensors in such kind of environments (e.g. LIDAR).

*A. Survey on Robustness*

In the front end/back end of a SLAM process, robustness problems related to inaccurate data affiliations can be dealt with. The front end has conventionally been given the right combination of results. The short-term data combination is more straightforward: when have a quick sampling frequency, monitoring the characteristics that correspond to the same 3-D mark was easy compared to robot dynamics. For instance, if a 3-D point across successive pictures, tracking was required and think the frame rate is high enough. Then it is guaranteed to have reliable tracking by standard Descriptor Matching approaches or Imaging flow. The camera sensor perspective does not shift dramatically at a high frame rate, so the characteristics at t+1 remain similar to those seen at t. A more difficult long-term application of data at the front end requires the identification and verification of loop closures. A Brute-force approach, which perceives current measuring characteristics (e.g. pictures) to match all recently detected characteristics, becomes unrealistic for loop closure detection at the front end. Models with bag-of-words [26] ensure that this traceability is reduced by discretising the storage and search efficiency. Bag-of-words should be organised into structured vocabulary trees [39] so that a wide-scale set of data can easily display.

Moreover, such strategies would not cope with severe glare alterations as visual sentences could no longer be paired. This resulted in the creation of alternative methods that take specific variants into account by coordinating sequences [13] or giving different views together in a single representation [49] or utilising data on both context and appearances [17]. Lowry et al. [49] provide a detailed analysis of visual location identification. Feature-based approaches are often included for laser-based loop closures for front ends of SLAM; for example, Tipaldi et al. [50] give 2-D laser-scanning applications.

The evaluation of the loop closure then involves more mathematical check-up steps to verify the value of the loop close easily. RANSAC is regularly considered for mathematical analysis and outlier exclusion in perception-based forms and links in it. In LIDAR methods, a loop closure can be verified by verifying whether the new laser scan corresponds to the original map that is, how tiny is the remainder of the test corresponding error.

The problem is multifold in dynamic environments. Next, modifications must be observed, detected, or tracked by the SLAM program. Although the conventional methods seek to remove the reactive aspect of the landscapes [18], reactive components are part of the prototype [ 12], [ 23]. The secondary task consists of forecasting guaranteed changes in the SLAM system and knowing how and when to modify the graph appropriately. Modern complex models with SLAMs either have multifaceted representations of the same position [16] or have a unified interpretation with a parameter that differs over time [14].

*B. Available Challenges*

This segment explores accessible problems and new SLAM development concerns.

*1) Error pruned SLAM with Recovery option:* While progress has been made on the SLAM history, in the case of exceptions, established SLAM algorithms remain fragile. The main reason for this is the fact that almost all efficient SLAM strategies are focused on iterative non-convex values. The outcome of the exclusion of a single outlier depends on what sort of the initial assumption fed to the optimisation; furthermore, the system is utterly vulnerable: it compromises the value of the calculation by including a single external component, which subsequently deteriorates the potential of apparent exceptions. Such types of shortcomings converge to a wrong linearising point from which recuperation, particularly in gradual setups, is not trivial. Failure secure and ineffective, i.e. the device needs to be aware of potential failures and provide restoration measures that could also repair the process correctly, should be an optimal SLAM approach. Any of the current SLAM approaches do not provide these capabilities. A closer alignment from the front to the back end is one conceivable way of achieving that, but the problem is still unanswered [42].

*2) Robustness to Hardware failure*: Such faults can be identified outside the range of SLAM when investigating device malfunction, which impacts the SLAM process, and therefore can play a vital role in control and cognition loss identification and mitigation. If the precision of a measurement deteriorates because of degradation, suitability issues to the surrounding or age, the reliability of the observations of the measurement is not in line with the back-end interference method, which leads to poor estimates. How to identify the behaviour of damaged sensors? How to improve the data of sensor interference accordingly? How can contradictory data from multiple sensors be addressed more broadly? In protection-critical uses (e.g. autonomous cars), this seems critical, when misreading of sensor data could endanger living creature [33].

*3) Relocalisation using loop closure:* Although the method could close loops in any time in the day or during different seasons, the resultant loop closure is quaternion in nature, based on appearances and as options for feature-based approaches. Feature-based methods remain the standard for compositional relocation, while component classifications lack underlying mathematical formalism to function properly under these conditions. Perception knowledge intrinsic to the SLAM issue could be used to solve such constraints, such as path comparison. A mapping can also be a useful addition, using one configuration of the sensor and a place in the same region, using a specific type of sensor [1].

*4) Drift in the generated Maps with time:* With a fixed and stable environment hypothesis in mind, the traditional Slam approaches were developed; but, both due to motion and the artefacts' intrinsic deformation, the natural world is non-rigid. An optimal SLAM system must be capable of comprehending and being able to produce maps of "any landscape" for conditions, like non-rigidity, over extended periods. Many attempts have been made to derive structure from unrigid artefacts from the computer vision group since the 80s. Latest performances for multi-rigid SfMs, such as [32], [37] are not very restrictive. The justification of smaller scale rebuilding has been discussed by Newcombe et al. [51].

*5) Automated tuning of the parameters:* In order to function correctly under a scenario, the SLAM systems need exhaustive parameter calibration. Such variables include levels that regulate similarity functions, RANSAC variables [27] and guidelines for planning whether new factors are applied to the maps or when an algorithm to look for similarities can be enabled. In entirely arbitrary circumstances SLAM must work "outside the box," techniques should be considered for automatically tuning the actively engaged parameters.

## IV. SLAM FOR AUTONOMY II: SCALABILITY

Though the most compelling example of recent SLAM implementations has been rendered in interior building settings, autonomous robots need to work for a more extended period in more extensive areas in many applications. Such technologies involve ocean inspection, non-stop maintenance of robots or substantial-scale comprehensive farming is continually changing cities. In this scenario, the continuous discovery of new areas and growing operating period, the scale of the parameter graph behind SLAM may expand indefinitely. In practice, the processing time and flash memory time are constrained by the robot's infrastructure [39]. SLAM techniques whose computing and memory sophistication persists constrained therefore are essential to model.

In the weirdest case, consecutive linearising approaches based on empirical deterministic algorithms suggest a quadratically increasing space usage. The resource usage tends to increase in the number of conditions linearly when exponential regression solvers [19] are used. It is exacerbated by the fact, that factor diagram scalability is less productive when exploring a location many times because edges or boundaries are introduced consistently into the same geographic region and thus undermine the map's sparse architecture.

This segment examines several established ways to control or at least to prevent the cause of a problem growth and explore outstanding issues.

*A. Survey on Scalability*

In this section, a study focused on specific ways to control the complexities of factor graph optimisation: 1) sparsification techniques, which sacrifice knowledge loss for memory and operational performance, and 2) Metrics that divide calculation into many robots and processors.

*1) Feature Sparsification – nodes and edges:* Scaling is accomplished in this class of strategies by reducing the number of nodes attached to the graph or by minimizing "informational" nodes and variables. Ila et al. [52] have an info-theoretical procedure to attach just extremely informative nodes and dimensions to the graph. Johannsson et al. [53] prevent the addition of new nodes to the graph if necessary, by creating new restrictions among the available nodes, whilst the number of factors only expands with the extent and not the scanning time of the explanation area. The information-based critique of Kretzschmar et al. [54] proposes which nudges to marginalize when improving the graph.

The cumulative path prediction would be another line of work which reduces the number of features measured over time that the opening SLAM method of this group would reflect the robot's persistent path with cubic splines [13]. Throughout their analysis, the nodes depicted in the variable graph are the trigger positions of the sliding pane. In the corresponding batch optimization proposal, Furgale et al. [55] suggested employing base features, in specific B-spline, to estimate pathway of robots.

*2) Parallel SLAM:* Parallel SLAM algorithms split calculations among different processors and shares the graph scalability workload. The main principle is to segment the map into multiple parts moreover to simplify the total diagram by interchanging local performance improvements and a worldwide enhancement of each subgraph. This idea goes back to the initial intent to solve substantial-scale maps [19] with the propose sub-mapping strategies to factor graph optimisation to arrange the submaps into a binary framework. They have been referred to as post-mapping methods.

*3) Distributed SLAM:* A method to map a broad ecosystem is to use several autonomous SLAM robots and to segment the

scene by divided spaces, one of which is controlled by another robot. This method has several vital variants: the centralized approach, during which robots generate sub-maps and pass regional details to the main station that carries out extrapolations [46], [49], and the decentralized version, whereby central data aggregation is not accessible.

*B. Available Challenges*

The history has significant shortcomings on specific facets of long-term activities, given the effort to control the sophistication of the variable graph refinement [27].

*1) Mapping in a large-scale area:* The dilemma of how the map can be stored for long-term purposes is entirely unexplored. Even when storage is not stringent, e.g. content is processed throughout the web, primitive interpretations as information-waste point maps, or conformal graphs, or the preservation of vision-based SLAM feature classifications is tedious. Specific approaches for a compact, established map [43] position and storage active deep restoration have been latterly suggested [36].

*2) Sparsification:* An underlying issue is how much data in the graph can be modified for long-term monitoring and how to determine when this knowledge is obsolete and could be simply ignored. If ever, when is it safe to forget? What could be overlooked, and what is important to keep? Is it possible to "take" portions of the graph and alert it if necessary? Although this obviously depends on the mission, in research, there was no reliable solution to these concerns [27].

*3) Multirobot applications towards robust mapping*: Whereas in the individual solutions to exceptional case refusal were suggested on the multi-robot SLAM applications. It is especially tricky for several causes to work with incorrect estimates. Firstly, the automatons may not share a common basis for comparison making it more complicated to identify and discard incorrect loop closures. Next, the automatons must find deviations of minimal and regional data in the clustered environment. A slightly earlier, try to address the above problem is [48], when robot cars consciously check observations of proximity by means of a rendezvous and docking policy before data is fused.

*4) Platforms with Limited Resources:* One entirely untouched question is how to apply the current SLAM methods in case of extreme technical limitations in the autonomous platforms. This is an essential issue in the reduction of the complexity of the system, e.g. handheld telephones, autonomous aerial vehicles or autonomous insects [29]. Most SLAM algorithms are quite complicated for those systems and optimisations should be used to set up a 'knob' that enables the precision to be carefully modified at the processing cost. Additional challenges emerge in the context of numerous robots: how should we maintain stable functionality of multi-robot squads while meeting limited bandwidth limits and connectivity trims.

## V. New Frontiers: Perception and Exploring

Intriguingly, the embedded sensor technology is the fundamental source of research in each SLAM solution. Appropriate algorithms are developed according to the characteristics of the sensors while considering the fusion of several sensors to mitigate fundamental errors in the individual sensors. In the beginning, distance sensors were used to get the aerial information such as acoustic and Light Detection and Ranging (LIDAR) sensors [23]. Such sensors have precise information about the depth but are not rich in features. Later systems have used mostly the vision sensors as the main source of information for perception such as monocular cameras and 360-degree cameras. Nevertheless, they lacked a comprehensive estimate of the depth. Then the more advanced type of sensors which could measure depth and colour are introduced such as RGB-D sensors and stereo cameras. They are used to generate point clouds while measuring the associated depth with the same point of reference. This section not only investigates all these sensors but reviews them for their efficacy based on their energy efficiency, scope, durability, maintenance, cost, preciseness, and spatial restrictions, which are essential for long-term deployments of autonomy.

Critical drivers for SLAM have been the introduction of new sensors and the use of new algorithmic and processing tools. This chapter discusses all traditional and innovative sensors as well as the challenges and opportunities they raise in the SLAM context Whereas the next paragraph addresses the position of deep learning as a significant field for SLAM, it analyses how this technology can boost, influence, or even summarise the question of SLAM.

*A. Innovative and Conventional Sensors for SLAM*

*1) Brief survey:* Traditional Sensors are reviewed in this section. Solution for any kind of a SLAM is solely depended on the used sensor technology.

*a) Ultrasonic SONAR Sensors:* Acoustic sensors are popular among early implementations of the SLAM solutions. Customized Echo-SLAM was implemented using microphone array and surrounding speakers as an Omni-directional acoustic sensor system [39]. This range only SLAM systems which function together with sensor networks should be used to make more feature-rich SLAM algorithms in the purpose of reducing the pose drift with time. Predominantly, these sensors are Sound Navigation and Ranging (SONAR) sensors which are using the time of flight strategies to compute the location of an obstacle. The Ultrasonic sensor, which is lying under the SONAR sensor category, is used extensively for robotics. Ultrasonic sensors are one the most inexpensive sensors used in SLAM implementations and hence popular among other acoustic

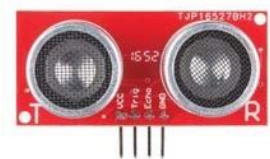

Fig. 3. HC-SR04 Ultrasonic Sensor [29]

sensors. They use ultrasonic waves (above 20kHz) which cannot be heard by humans.

These are appropriate for dark conditions due to insusceptibility towards illumination and opacity. However, these sensors operate well in the presence of dirt and humidity, but huge contaminations can influence sensor readings [44]. Moreover, because the sound waves require a transverse medium, they do not work in a vacuum. Also, these sensors are vulnerable to the reflection distortion with the surface smoothness since smooth surfaces tend to absorb acoustic waves rather than reflecting them. The dimensions of Acoustic sensors are generally limited to few inches due to its compact form and the power requirements for an acoustic sensor is ranges from milliwatts to watts. Considering all of these facts, Ultrasonic sensors are remain as an affordable and robust distance measuring equipment given that the accuracy of the maximum depth range of these sensors is about 1% to 3% whereas the environment temperature, moisture level and air pressure for which compensation techniques are usually employed with the sensor itself [38].

*b) LIDAR:* It is the abbreviation for Light Detection and Ranging sensor. Since the 1960s, LIDAR's fundamental architecture has been around for obtaining aerial distance mapping. LIDAR's job is like an ultrasonic detector. LIDAR utilises electromagnetic waves in the spectrum of visible light as a radiation reference instead of using sound waves. By activating to 1,000,000 bursts per second, a LIDAR produces a 3D Visual representation of its environment known as the Point Cloud. In determining the depth analysis, LIDARs can provide 360 degrees of perception and are quite precise (~±2 cm) [27].

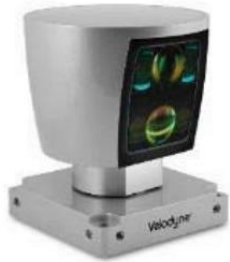

Fig. 4. Velodyne HDL-64E. Popular sensor among self-driving cars [29].

*c) Depth sensing cameras:* Light-emitting distance cameras were not new devices; however, with the introduction of the Xbox Kinect game console, they became mainstream equipment. They function on various principles, such as structured light, the flight of time, diffraction gratings, or shutter speed coded. Structure-light cameras operate by triangulation; the range between both the cameras and the sequence projector, therefore, limits their precision [12], [23]. From the other hand, the precision of Time-of-Flight (ToF) sensors relies just on the TOF sensor, thereby having the maximum precision distance. Whereas a poor signal-to-noise (SNR) ratio and high cost marked the very first generation of ToF and structured-light cameras, they quickly grew famous with computer game technologies, which helped make them cheaper and enhance their precision. Although field cameras bring their own source of light, they also operate in darkened and untextured environments, allowing impressive SLAM performance to be achieved [18].

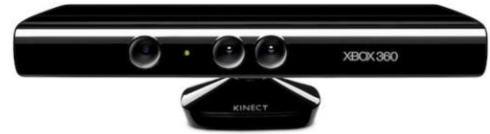

Fig. 5. Kinect XBOX 360: RGB-D Sensor [29].

*d) Event-based cameras:* In contrast to conventional framing-based cameras that send full frame rates, event-based cameras like the dynamic vision sensor [15] or the Asynchronous Time-based photograph sensors [16], only addresses the local bitmap-level motion alterations to an incident when they take place.

These have five key benefits relative to traditional frame-based cameras: 1 ms time delay, up to 1 MHz refresh frequency, a dynamic contrast of up to 140 dB (up from 60–70 dB of regular camera systems), 20 mW throughput (up from 1,5 W of conventional camera systems), and minimal bandwidth and processing demands (because only shifts in brightness are broadcasted). This allows for the construction of a new category of SLAM methodologies, which can rely on high-speed movement scenes [9] and steep-dynamic range [12][27], in which traditional cameras are unsuccessful. As the throughput is a set of unpredictable activities, though, standard computer-vision techniques focused on structures are not appropriate. This calls for a radical shift over the last 5 decades from standard computer perception strategies. Recently, event-based techniques for the location of events and visualization have been proposed [ 32], [ 47]. The development purpose of these kinds of algorithms is to enable every arriving event to alter the entire system approximate state dynamically, thereby retaining the sensor's eventual existence and enabling microsecond-latency algorithms to be designed [178].

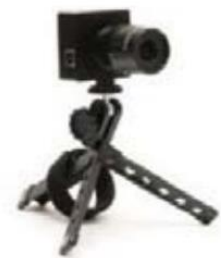

Fig. 6. DVS128 Event Camera [29].

*2) Open Problems:* Effective range and intervention with other additional ambient light sources (such as daylight) is the main

drawback of active range cameras; these deficiencies can be enhanced by more power.

Light field sensors in SLAM were hardly ever utilized since the quantity of data generated, and more processing power is required. Nonetheless, current researches have demonstrated that it is ideal for SLAM implementations since it allows for linear optimization of a movement approximation dilemma and could provide more precise motion predictions if appropriately designed [49].

What's the perfect SLAM sensor? Of course, there is one question: how is it to conduct coming decades, long-term SLAM research using succeeding sensor systems? The quality of a given pair of algorithm sensors for SLAM obviously relies both on the transducer constraints, on the algorithm, and on the surrounding [22]. There hasn't yet been a comprehensive analysis of the selection of architectures and sensors for better results. Research by Censi et al. [56] showed that effectiveness for an application also hinges on the measuring strength. It also implies that several sensors may be automatically turned on and off to the desired quality level for the optimum sensing system or calculate the same behaviour by various physical standards of reliability [40].

*B. Deep Learning approach*

A paper that aims to take longer-term directions in the SLAM without considering deep learning should be a mistake. It has transformed image processing and already makes significant additions to conventional robotic systems, along with SLAM implementations [22], [51].

Research scientists have already demonstrated that a deep neural network can be trained to reconstruct the interframe position among two images captured from a travelling robot immediately from the initial object pair [53], permanently eliminating the traditional geometry of graphical odometry. Additionally, the 6DoF can be found in the regression forest [47] and with the deep convolutional neural network [29], and a single frame can be used to determine the depth of an environment [28],[41] and [58].

## VI. CONCLUSIONS

In the last 30 years, tremendous development on simultaneous localization and mapping has happened [27]. Various fundamental problems were resolved across the way, with the implementation of new technologies, new instruments and modern cognitive tools that introduced numerous new and exciting questions.

Whereas finding an answer to the question that "Is SLAM required?" That answer depends on the request, but often the reply is a definite yes [27]. In a wide range of real-world contexts, from autonomous cars to mobiles devices, SLAM and associated technologies, such as visual-inertial odometry, are rapidly being used. For cases where telecommunication options, such as GPS, have not been accessible or available with a lack of precision, SLAM strategies are commonly used to deliver accurate metric positioning. It is possible to imagine that cloud-based positioning and navigation applications coming online and maps being commoditized because of the popularity of mobile phones and positioning data [9].

Precise localization is often done in some systems; for instance, self-driving cars are linking established sensor data to an advanced high-definition map of the scene [54]. The online SLAM would not be essential if the a priori mapping is precise. Nevertheless, interactive online map updates are needed to deal with development or significant improvements in road systems for operations in increasingly complex ecosystems. The distributed refreshing and construction of spatial maps generated by large autonomous fleets is an important area for future research [51].

There will be tasks that are more suitable to distinct SLAM flavours than others can be identified. For example, a spatial map can be used to evaluate the accessibility of a particular location, but is not ideal for movement planning and minimal control; a locally compatible map is well optimized for avoiding obstacles and local encounters with the terrain, but may compromise accuracy; a consistent map enables the robot to navigate broad routes [12].

One may also formulate instances in which SLAM is entirely unnecessary and can be supplemented by other methods such as visual command servoing, or "educating and recurring" to carry out ongoing navigation tasks. The more general way to pick the best SLAM model is to imagine of SLAM as a framework for calculating adequate statistics to reiterate all the robot's observed data and in that context which data is fiercely task-dependent to retain.

With respect to the common problem "SLAM is achieved?" This paper review that the autonomous robot, ecosystem and performance combination cannot be addressed until we reach the robust perception era. Significant difficulties and critical concerns stay open for several implementations and ecosystems [34]. Further work in SLAM is required to achieve more reliable perception and navigation for durable autonomous systems. SLAM is not entirely solved as an academic attempt with significant real-world impacts [35].

Unsolved issues cover four main aspects: overall performance, wide-level awareness, resource consciousness, and process-driven conclusion. The layout of a SLAM auto-tuning is a significant difficulty in terms of robustness, with many aspects widely not investigated [24]. The durability of autonomy requires a significant quantity of scientific research on strategies for generating and sustaining maps as well as rules that identify when remembering, updating, or forgetting information; similar difficulties arise in robotic structures which are severely restricted to resources [50], [52].

In addition to addressing many achievements and challenges ahead for the SLAM community, opportunities are explored related to the use of new sensor data, new techniques (e.g. convex relaxation and duality theory, or deep learning), and the role of active detecting [27]. SLAM is still an invaluable cornerstone for most robotics implementations and, despite incredible progress over the past decades, existing SLAM technologies are still far from offering intelligent, substantive, and durable environmental designs analogous to those generated and used seamlessly by humans.